\documentclass[usenames,dvipsnames,journal]{IEEEtran}
\usepackage{pgfplots}
\usepackage{lipsum}
\usepackage{multicol}
\usepackage{cite}
\usepackage{url}
\usepackage{textcomp}
\def\BibTeX{{\rm B\kern-.05em{\sc i\kern-.025em b}\kern-.08em
T\kern-.1667em\lower.7ex\hbox{E}\kern-.125emX}}
\usepackage [english]{babel}
\usepackage{color, colortbl}
\usepackage{amsmath}
\usepackage{amssymb}
\DeclareMathOperator{\E}{\mathbb{E}}
\usepackage{algorithmic}
\usepackage[english]{babel}
\usepackage[utf8]{inputenc}
\usepackage{algorithm}
\usepackage{array}
\usepackage[caption=false,font=footnotesize]{subfig}
\usepackage{caption}
\usepackage{fixltx2e}
\usepackage{stfloats}
\usepackage{url}
\usepackage{eqparbox}
\usepackage{arydshln}
\usepackage{mwe}
\usepackage{caption}

\usepackage{subfig}

\usepackage{blindtext}
\title{here title}
\let\oldtwocolumn\twocolumn
\renewcommand\twocolumn[1][]{%
    \oldtwocolumn[{#1}{
    \begin{center}
           \includegraphics[width=\textwidth]{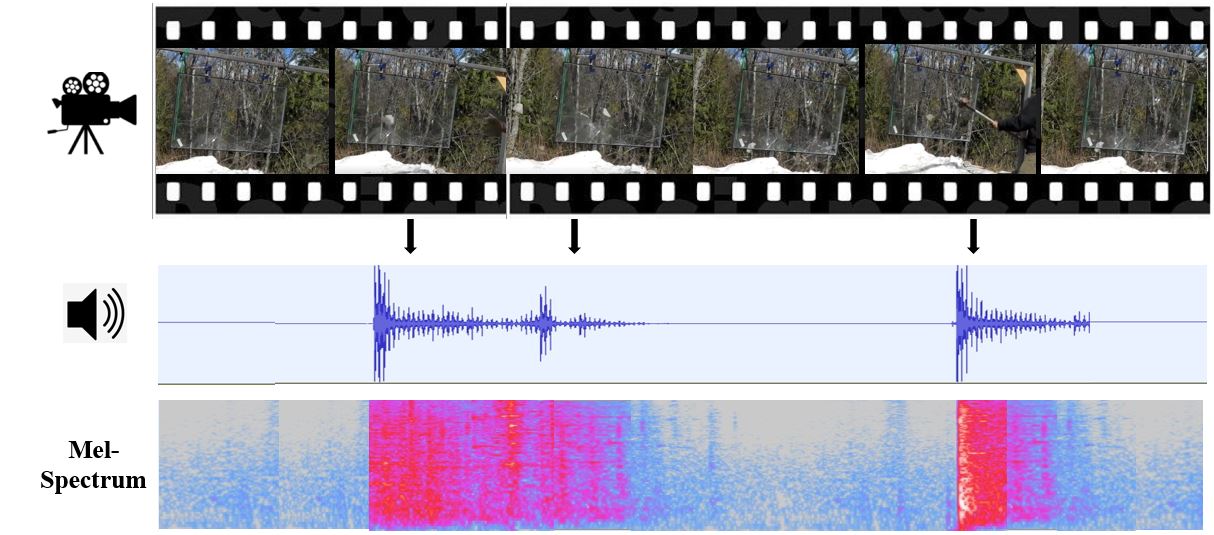}
           \captionof{figure}{Problem description: Visually synchronous sound synthesis capturing temporal action information. Our proposed model locates the temporal action changes in subsesquent frames of a video and generates the sound accordingly.}
           \label{fig:fig1}
        \end{center}
    }]
}

\DeclareUnicodeCharacter{2212}{-}
\begin{document}
\captionsetup[figure]{labelfont={bf},labelformat={default},labelsep=period,name={Fig.}}
\captionsetup[table]{justification=centering, labelsep=newline}
\title{\title{FoleyGAN: Visually Guided Generative Adversarial Network-Based Synchronous Sound Generation in Silent Videos}}

\author{Sanchita~Ghose,~\IEEEmembership{Student Member,~IEEE} and ~John~J.~Prevost*,~\IEEEmembership{Senior Member,~IEEE}
\thanks{S. Ghose (sanchita.ghose@my.utsa.edu) and J. J. Prevost (jeff.prevost@utsa.edu) are with the Department of Electrical and Computer Engineering, The University of Texas at San Antonio, One UTSA Circle, San Antonio, TX 78249 USA.\newline
*- Indicates the Corresponding Author\newline
This research is supported by the Open Cloud Institute at UTSA.
}}

\maketitle

\begin{abstract}
Deep learning based visual to sound generation systems essentially need to be  developed particularly considering the synchronicity aspects of visual and audio features with time. In this research we introduce a novel task of guiding a class conditioned generative adversarial network with the temporal visual information of a video input for visual to sound generation task adapting the synchronicity traits between audio-visual modalities. Our proposed FoleyGAN model is capable of conditioning action sequences of visual events leading towards generating visually aligned realistic sound tracks. We expand our previously proposed Automatic Foley dataset to train with FoleyGAN and evaluate our synthesized sound through human survey that shows noteworthy (on average 81\%) audio-visual synchronicity performance. Our approach also outperforms in statistical experiments compared with other baseline models and audio-visual datasets.

\end{abstract}

\IEEEpeerreviewmaketitle

\maketitle

\section{Introduction}

\IEEEPARstart{F}{oley} recording, an inevitable part of film production, not only mimics the sound that an actor is doing on the screen, but also provides an added realism and clarity to the scene. 

Today's film production teams are mostly dependant on Foley tracks for those movie scenes where background sound is not present at all or the original recording does not come through acceptably well. In these situations they either look for available recorded foley tracks of respective class or prepare studio setup for mimicking the sound through Foley artists and recording it with the screenplay in a noise-free environment. Since the later option certainly turns out to be more pricey, in most cases filmmakers prefer to get available recorded tracks from online or some other sources at low cost. Apparently this seems like an easy solution but they often encounter lack of synchronicity between the video and the overlaying sound. Here comes the necessity of applying deep learning algorithm that can learn the correspondence between audio and video signal and can generate the sound accordingly for the given video clip. In our previous work \cite{9126216}, for the first time we addressed the traditional foley generation problems and proposed two deep learning models for automatic foley generation. While making a sound of Foley, we specially try to enhance the properties that are intrinsic in the video file, the director is asking to create a Foley sound that an audience person will instantly associate  with the relevant video. It is about sensory augmentation and we want the audience to be more engaged than normal. This was our initial approach to this crossmodal problem, that needs to be paid more attention in the time synchronization domain. 

In this paper, we propose a visually guided class conditioned deep adversarial Foley generation network called "FoleyGAN" where we present as an advancement in automatic foley synthesis deep neural network from silent video clip. Since, sound plays a crucial role to perceive the inherent action information of most of the visual scenarios of real world and auditory guidance can assist a person or a device to analyze the surrounding events more effectively, our proposed network has also the potential to serve as an IoT (Internet of Things) system which is able to learn the correspondence between visual and audio modalities along with synthesizing synchronous sound tracks from visual signals.

Generative Adversarial Networks (GANs) \cite{goodfellow2014generative} has started immensely favoring the researchers as a promising deep generating model particularly for high quality image generation applications (e.g. \cite{donahue2019large}, \cite{karras2019style},\cite{dumoulin2016adversarially,donahue2016adversarial,karras2020analyzing}).  
Notable advances are found in utilizing (GANs) for audio and music generation \cite{donahue2018synthesizing}, \cite{vasquez2019melnet}, \cite{donahue2018adversarial}, \cite{engel2019gansynth}, \cite{yang2017midinet},\cite{chen2017deep} as well though adversarial audio generation still remains a highly challenging task because of some intrinsic differences between sound waveforms and image signals. Sound waves generally show higher periodicities than image signals which leads to use more sophisticated filters with large receptive fields while processing them. In addition, generated audios are more likely to be affected by annoying "checkerboard" artifacts those can be easily avoided in a generated image using GAN. Recent researches work with spectral representations of audio for adversarial generation. However, none of these  approaches have considered about time-action synchronicity traits as a visual guidance to condition the sound generator of GAN along with sound class information, which is infact the key aspect of our "FoleyGAN" network (Fig.1). On top of that, we incorporate efficiently scaled-up (512 $\times$ 512) BigGAN \cite{brock2018large} architecture as our base generative network that benefits us greatly to synthesize high resolution spectrogram generation invertable to sound track via ISTFT \cite{crochiere1980weighted}. In addition to using latent space and class information as inputs we condition the BigGAN generator with visual guidance. Furthermore, we expand our previously proposed "AFD" dataset \cite{9126216} and the discriminator network is pretrained with the spectrogram images of soundfiles of the updated dataset to differentiate between generated and actual samples.

\graphicspath{ {Figure} }
\begin{figure}[h]
\includegraphics[width=85mm]{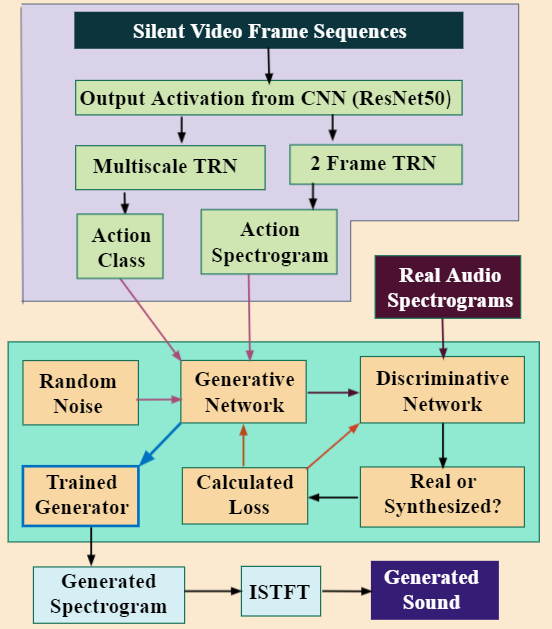}
\caption{FoleyGAN Model: the upper section utilizes TRN models for predicting class and temporal action information that are passed to the lower section's GAN structure as guidance to generate spectrogram from random noise. Lastly generated spectrograms are converted to sound via ISTFT.}
\end{figure}

The Fig.2 shows the proposed FoleyGAN network are consists of two major neural network blocks: video action recognition network (upper block consisting CNN and TRN \cite{zhou2018temporal} architectures) followed up with visually guided class conditioned GAN network (lower block) for sound generation. The first block provides the prediction of the action category of the respective input video as well as prediction weights of the action occurrences over the video time duration from which we are generating action spectrogram. These two outputs are forwarded to our next sound generative network using GAN principle. Finally, the generated spectrogram is inverted via ISTFT to obtain the visually synced sound track for the respected video clip.

Previously in AutoFoley \cite{9126216} we proposed two separate deep neural networks (e.g. Frame Sequence and Frame Relation Networks) for predicting the existing action in the video frames. Since, the overall performance of both models are found quite similar, we can opt any of these models for the later expansion of this research. However in this work, we have to impose our higher concern on reducing the computational complexity as we are here integrating a scaled up GAN architecture (e.g. BigGAN) for high resolution spectrogram generation. Besides, in this work we are aiming to advance the earlier proposed automatic sound synthesis system with time synchronicity features. Therefore, we intentionally pick the Frame Relation Network that is not only capable of capturing the temporal relations between two consecutive video frames leading to predicting the action happening in the scene, but also uses limited video frames as inputs that are fed into a more simpler multilayer perceptron (MLP) structure. In addition, we are able to condition our generator network with the relational reasoning information between two sequential frames with the help of temporal relation statistics.

The significant contributions made by this paper are: 
\begin{itemize}
\item We take the initial step toward automatic Foley generation in a silent video clip using visually guided class conditioned generative adversarial network, taking into consideration of the time-action synchronicity requirement in the highly diverse "movie sound effects" domain.
\item We introduce a concept of conditioning the generated samples of a GAN network with temporal visual information of a video frame sequence that can be deployed for automatic Foley synthesis as well as other multimodal applications.
\item We expand our previously proposed "Automatic Foley Dataset (AFD)" for efficient training purpose.
\item We present image generating BigGAN architecture trained on "AFD" dataset for realistic and synchronous sound synthesis of 3 second duration for a rarely addressed multimedia application field.
\item For the performance analysis of our generated sounds, we perform qualitative, numerical experiments and conduct a human survey on our generated sound quality as well as video with sound alignment ability in respective visual events. 
\end{itemize}

The paper is structured as follows, in section II and III, we present related works and brief review of GAN background. In section IV, we describe our detailed methodology (audio and video preprocessing steps, a video action recognition followed by a sound generation network) and the complete algorithm. Sections V and VI, provide the explanation of our extended AutoFoley dataset, training details with specifications on hyperparameter tuning, along with model evaluation result analysis through numerical, qualitative and ablation experiments to asses the overall performance respectively. Finally, section VII concludes with summarizing substantial points over and above future directions of this work.

\section{Related Work}

\subsection{Foley Generation}
Automatic sound effect creation from 3D models has been approached in \cite{van2001foleyautomatic} through dynamic simulation and user interaction
In our recent work \cite{9126216} deep learning is deployed in the application of automatic Foley generation for the first time, where we propose a deep learning solution to predict sound in silent video clips of movie scenes and then synthesize Foley from the predicted features. In this paper, we utilize conditional generative adversarial training on our predicted video categories to generate foley of that respected class.

\subsection{Audio-Visual Correlation}

We observe impactful audio-visual events happening around us where sounds play a vital role representing the event status. Human correlation capability in perceiving these two modalities simultaneously lead them to act accordingly in their real life events. Taking inspiration of this fact, \cite{arandjelovic2017look, owens2016ambient, aytar2016soundnet, wang2016visualizing, 9126216} utilizes these audio-video correspondence properties for training their neural networks with unlabeled video data. The audio-visual relationship is employed to develop deep neural network in various fields of applications e.g. for the  material recognition task \cite{owens2016visually}, sound source localization task in video \cite{arandjelovic2017look,gaver1993world, kingma2014adam, majdak20103, shelton1980influence, bolia1999aurally, perrott1996aurally,aytar2016soundnet}, audio source separation tasks \cite{owens2018learning}, audio event identification task for video analysis \cite{takahashi2017aenet}, video action recognition to automatic foley generation task \cite{9126216}. Likewise, advanced research approaches are proposed in \cite{majdak20103, bolia1999aurally, perrott1996aurally} has provided  on localizing a sound source against visual data in 3D space utilizing human adaptation ability to observe audio-visual events. In \cite{wang2016visualizing}, an automatic video sound recognition and visualization framework is proposed, where nonverbal sounds in a video are automatically converted into animated sound words and are placed close to the sound source of that video for visualization. In addition, attention mechanism learning network for the sound source proposed in \cite{xu2015show}, semantic guided modules (SGMs) performed in \cite{yu2019weakly} for action recognition to extract spatial-temporal features from videos show promising applicability in audio-visual association properties. We are highly motivated by these researches on audio-visual relevance, hence in this work we aim for improved mapping of audio-video features by expanding our AutoFoley deep neural network with an efficient generative adversarial model.

\subsection{Sound Synthesis from Videos}

Understanding the synchronizing capability of human brain for audio and video modalities simultaneously, \cite{owens2016visually, zhou2017visual, zhangvisually, chen2017deep, 9126216, gao20192, zhou2019vision,chen2020generating, liu2021towards} propose different neural networks for sound synthesis from visual inputs. Research in \cite{morgado2018self} audio generation for the full viewing sphere when 360◦ video and corresponding mono audio are given, whereas in their later work \cite{gao20192}, they leverage object configurations in videos for transforming mono channel to binaural audio. Similar video-based audio spatialization research is shown in [\cite{li2018scene}. Prior work in \cite{zhou2017visual} shows natural sound generation from videos captured in the wild whereas the AutoFoley framework \cite{9126216} synthesize Foley tracks in silent video frames. Another approach for sound generation from visual inputs is presented in \cite{chen2017deep} using conditional generative adversarial networks. Recent work in \cite{chen2020generating} proposed a spectrogram based sound generation model named REGNET where authors introduced audio forwarding regularizer to pass the missing information while training. In this work, we develop deep learning model comprising visual action recognition and adversarial audio synthesis network to generate realistic Foley tracks for silent movie clips. 

\subsection{Audio Generation with GAN}

GAN extensive potentials in computer vision and image generation field (e.g. \cite{donahue2019large}, \cite{karras2019style}, \cite{dumoulin2016adversarially,donahue2016adversarial,karras2020analyzing}) massively encourage researchers  to deploy the principles in audio generation domain as well. Being inspired by image inpainting recently authors in \cite{zhou2019vision} perform audio inpainting as form of spectrograms with GAN. Earlier works in \cite{donahue2018synthesizing}, \cite{vasquez2019melnet}, \cite{donahue2018adversarial}, \cite{engel2019gansynth}, \cite{yang2017midinet},\cite{chen2017deep} show a clear direction of using generative adversarial training with audio signals. However \cite{engel2019gansynth}, [\cite{donahue2018adversarial} portray the challenges to train GAN with audio waveform compared to image matrices. Therefore, spectral representations of audios are preferred while training adversarial audio generation. The phase-gradient heap
integration (PGHI) \cite{pruuvsa2013large} algorithm proposed in TiFGAN
paper \cite{marafioti2019adversarial}, represents an improved reconstruction technique
of the audio from the spectrogram with minimal loss. Authors in \cite{marafioti2019adversarial}, trained GAN on short-time Fourier features to mitigate the problems of generating audio in the short-time Fourier domain. In our previous work \cite{ghose2020enabling} for the first we propose a System of Systems framework of audio generation for visual inputs exploiting BigGAN \cite{brock2018large}. Recently, authors in \cite{haque2020high} utilizes BigGAN architecture for adversarial audio generation in guided manner. Our proposed FoleyGAN architecture is a noble approach to apply BigGAN in movie sound production domain where we are synthesizing the audio for silent movie clips taking visual guidance.

\section{Generative Adversarial Network (GAN) Basics}

Generative Adversarial Networks (GANs) proposed in \cite{goodfellow2014generative} includes a generator network $G$ and a discriminator network $D$ taking part in a min-max game where the two networks play in adversarial manner throughout the training process. The training objective of G network is to map random vector $z\in Z$ into generated samples by minimizing the following value function (Eq 1)  whereas the D network, that judges between real and generated examples is trained to maximize the value function. Here $z$ belongs to random noise distribution $p_z$ and $p_{data}$ denotes the target data distribution.

\begin{equation}
\begin{split}
    \min_G\max_D{V}(D,G)=\E_{x\sim{p_{data}(x)}}[\log{D(x)}] +\\ \E_{x\sim{p_{z}(z)}}[\log{(1-D(G(z))}]
\end{split}
\end{equation}

In conditional GANs approach (Equation 2), conditional information (e.g. labels of images) are passed to the generator and discriminator networks where $y$ represents the condition variable. 

\begin{equation}
\begin{split}
    \min_G\max_D{V}(D,G)=\E_{x\sim{p_{data}(x)}}[\log{D(x|y)}] +\\ \E_{x\sim{p_{z}(z)}}[\log{(1-D(G(z|y))}]
\end{split}
\end{equation}

In this work, we use hinge loss for updating the generator and the discriminator in our visually guided sound generation network. In a conditional GAN network, hinge loss for the discriminator and generator are calculated as,
\begin{equation}
\begin{split}
L_D= L_{Dreal} + L_{Dfake}\\
=\E_{(x,y)\sim{p_{data}}}[\max(0,1-D(x,y))] +\\ \E_{z\sim{p_{z}},\sim{p_(z)},y\sim{p_{data}}}[\max(0,1+D(G(z,y),y))]
\\
L_G=-\E_{z\sim{p_{z}},y\sim{p_{data}}}[D(G(z,y),y)]
\end{split}
\end{equation}

\section{Proposed Research Method}
We section our proposed architecture into two networks: i) video action recognition network and ii) sound generation network. We explain these network details in the following  subsections. The graphical representation of the complete FoleyGAN architecture is presented in Fig.3.

\begin{center}
\graphicspath{ {Figure} }
\begin{figure*}[ht]
\includegraphics[width=180mm]{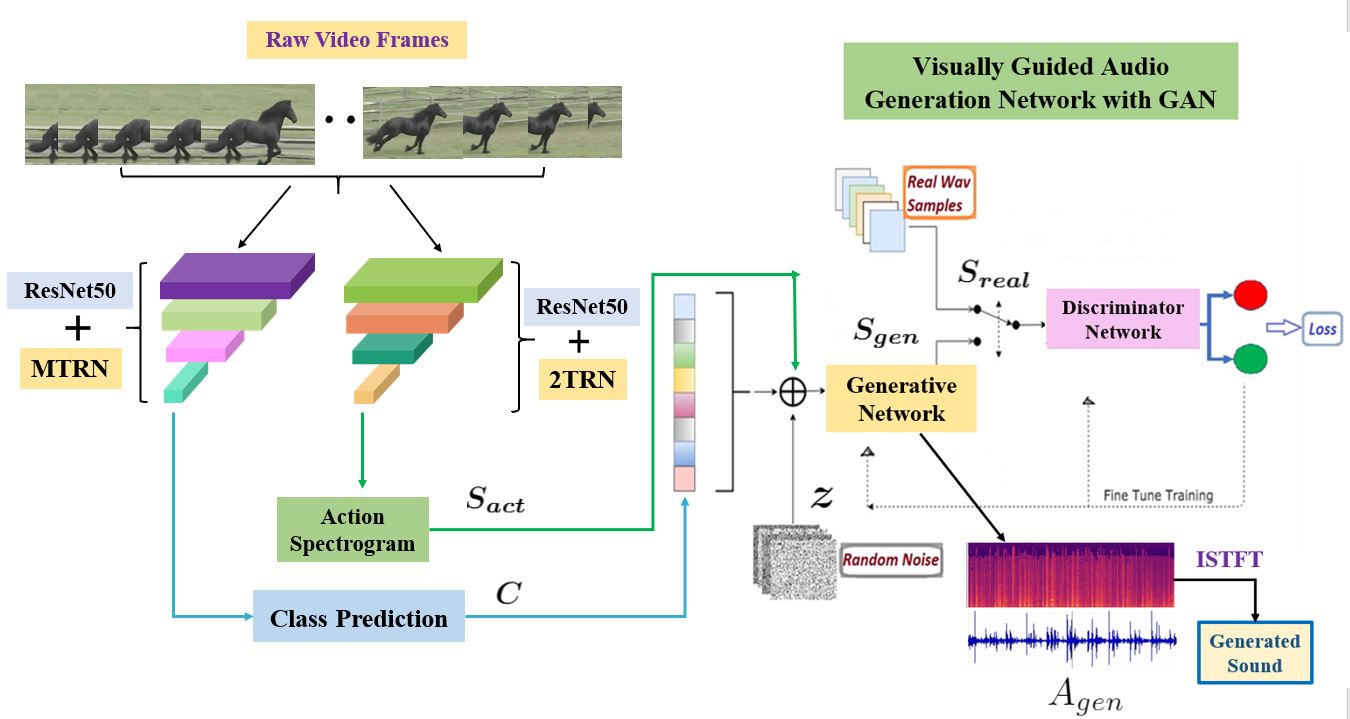}
\caption{Proposed Sound Generation Architecture in detail.}
\end{figure*}
\end{center}

\subsection{Video Action Recognition Network}
We pick the frame relation model from \cite{9126216} for class prediction because of its excellent performance on learning temporal dependencies from visual frames than the other prediction model with less computational complexity. The video action recognition network provide the prediction of the overall action category along with the frame by frame identical action occurrence probabilities of the entire video clip exploiting the multiscale and 2-frame temporal relational networks \cite{zhou2018temporal} principle respectively. The detailed methods are explained in following paragraphs.

\subsubsection{Video Action Class Prediction}

We use a fused network comprised of CNN and multiscale temporal relation network (TRN), proposed in \cite{zhou2018temporal} to identify the action occuring throughout the video clip. Here, we compute the temporal relation composite functions $\boldsymbol{R}_Q$ using the following equation where $Q =[2,3,...8]$ represents the number of video frames under consideration:

\begin{equation}
\begin{split}
&\boldsymbol{R}_2 = \boldsymbol{h}_\phi(\sum_{j<k}\boldsymbol{g}_\theta(\boldsymbol{F}_j,\boldsymbol{F}_k))\\
&\boldsymbol{R}_3 = \boldsymbol{h}'_\phi(\sum_{j<k<l}\boldsymbol{g}'_\theta(\boldsymbol{F}_j,\boldsymbol{F}_k,\boldsymbol{F}_l))\\
\end{split}
\end{equation}

Here, $\boldsymbol{F}_j$, $\boldsymbol{F}_k$, $\boldsymbol{F}_l$ represents the activation output obtained from the pretrained ResNet-50 \cite{He_2016_CVPR} CNN architecture at $j^{th}$, $k^{th}$, $l^{th}$ frame of the video. We train the ResNet-50 model with the $n$ number of soundless video frames $[I_1, I_2,....I_n]$ of each video ($\boldsymbol{V}$) from our train dataset. In this equation, ${\boldsymbol{h}}_\phi$ is a single layer
and ${\boldsymbol{g}}_\theta$ is a double layer multilayer perceptron (MLP) associated with 256 units per layer. These functions compile features of video frames at different temporal order and are unique for each ${R}(\boldsymbol{V})$. In this way we calculate the composite temporal function over time among up to 8 frames as $\boldsymbol{R}_8(\boldsymbol{V})$ since upto this frame number we find optimum result through ablation studies on TRN network in our earlier work \cite{9126216}. Finally, we sum all the temporal relation functions (equation 2) to compute the action category $\boldsymbol{C(\boldsymbol{V})}$ happening in the entire video clip.

\begin{equation}
\boldsymbol{C(\boldsymbol{V})} = \boldsymbol{R}_2(\boldsymbol{V}) + \boldsymbol{R}_3(\boldsymbol{V}) +...+\boldsymbol{R}_8(\boldsymbol{V})
\end{equation}

\subsubsection{Video Action Spectrogram Generation}

Our intuition is to obtain the relational reasoning information between  two  sequential  frames over the complete time duration of the video with  the  help  of  temporal relation equation of $R_2$. We plot these values over time to get a time-series graph representing the probabilities $\boldsymbol{P_{act}}$ of similar action occurrence of two subsequent frames over the whole time period of the video. Next, we convert this time series plot into spectral representation by computing STFT and reshape it into (512 x 512 x 3) dimension. Therefore, we get a 3D matrix, $\boldsymbol{S_{act}}$ that we name as video action spectrogram since it contains the frame by frame similar action occurrence probabilities of each video (Fig.4). We are going to condition  our sound generation network with this visual guidance to sustain the temporal synchronicity between audio and visual inputs.

\begin{center}
\graphicspath{ {Figure} }
\begin{figure}[t] 
\fcolorbox{black}{white}{\includegraphics[width=83mm]{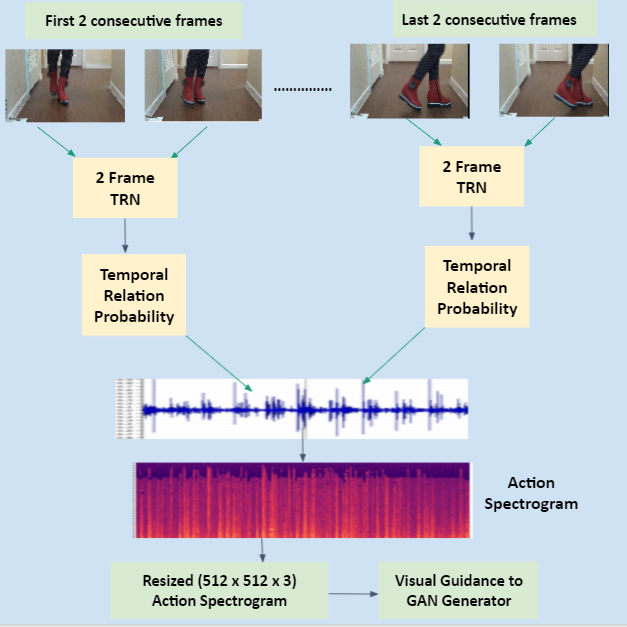}}
\caption{Action spectrogram formation for visual guidance for audio generation with GAN generator network.}
\end{figure}
\end{center}

\subsection{Sound generation network}

\subsubsection{Preprocessing of Sound Data for Training}

In objective to train our generative model (image generating GAN) with sound data, we have to represent our sound files as three dimensional matrix without losing  magnitude and phase information contained with individual tracks. For this at first, we extract audio from video recordings and clip them into 3 sec duration. We convert audio files into mono-wave files and then compute their spectrograms by calculating STFT with the help of TensorFlow's built-in functions. We use Hanning window and sample frequency of 44kHz. We select stride of 256 and frame size of 1024 allowing windows to overlap 75${\%}$ with 513 frequency bins. In order to obtain a three dimensional image like matrix, we use padding in time axis. Finally, our complex spectrogram of each sound file become a (512,512,3) matrix containing both the magnitude and phase information of the original audio in the 1st and 2nd channel respectively. For the 3rd channel we again apply zero padding that we extract later through depadding during the reconstruction process. Finally, we prepare the sound spectrogram features appying a mel-filter
bank to convert the frequency scale into the mel-scale. Since our generator network applies tanh nonlinearity function, we scale the log magnitudes and phase angles within -1 to 1 range to comply with the generator model.

\subsubsection{Generation of Visually Guided Sound}
Likewise SpecGAN model proposed in \cite{donahue2018adversarial}, our deep sound generation network is basically a frequency-domain sound synthesis GAN architecture. The proposed generation network is trained with the spectrogram inputs by performing short-time Fourier Transform (STFT) \cite{allen1977short} on the audio samples. The generated output spectrograms are inverted using (ISTFT) method \cite{crochiere1980weighted}. The objective of feeding spectrogram inputs to the generation network is to deploy the proficiency of GAN in high resolution image generation tasks. In this proposed model, we adopt BigGAN \cite{brock2018large} for adverserial sound synthesis by generating high fidelity spectrogram images of multiple categories through large scale GAN training. The generator and discriminator network follows BigGAN (512 × 512) image generation architecture capable for generating high resolution spectrogram images of multiple sound classes. 

In brief, BigGAN is a high resolution and high fidelity class-conditional image generating GAN model that significantly improves the inception score using higher batch sizes with increased width in each layer. Being a class conditional GAN, it takes image class information and a point from latent space as input. Rather than using the pretrained weights of BigGAN trained on natural images  from ImageNet dataset, we train the model with our generated spectrograms to follow our goal for adversarial sound synthesis. As previously mentioned, the class output, $\boldsymbol{C}$ resulted from the prediction network and the action spectrogram, $\boldsymbol{S_{act}}$ are fed into the generator. Being conditioned by the video action information, the generative network produces spectrogram of the predicted class taking some random noise, $\boldsymbol{z}$ as input. Next, the generated image, $\boldsymbol{S_{gen}}$  is passed to the discriminator block pretrained with original spectrogram image $\boldsymbol{S_{real}}$ of that predicted class. The discriminator network distinguish between the real $\boldsymbol{S_{real}}$ and synthesized spectrogram $\boldsymbol{S_{gen}}$. Likewise BigGAN, we adopt orthogonal regularization technique and truncation trick to boost the performance and improve generated spectrogram quality. With “truncation trick” our generator takes less random numbers while generation leading towards to output more realistic images.  Finally $L_D$ and $L_G$ losses are calculated ( as Equation 5) and fed back to generator and discriminator blocks to update their weights at end of each training epoch.

As the training proceeds, the generator gets closer to synthesize spectrogram that misguides the discriminator identifying the differences between original and generated images. At the end of the training the generator learns the pattern and features of the original spectrograms and generate representative spectrogram images of 512 × 512 resolution classified as real by the discriminator. For the complete architecture and parameter details of BigGAN's generator and discriminator blocks we direct readers to the appendix section of the original paper \cite{brock2018large}.

\begin{algorithm}[H]
\caption{Visually Guided Adversarial Foley Generation}

\textbf{Input: }Silent video frames ($I_1, I_2,...I
_N$), training audio\\tracks ($A_1, A_2,...A_N$) and random noise $z$.\\
\textbf{Output: }Generated audio tracks ($A_{gen}$).

\begin{algorithmic}[1]

\STATE{ $V_t \gets CNN(I_N)$}
\STATE{$Prob_{class} \gets MTRN (V_t)$}
\STATE{$Prob_{seq} \gets 2TRN (V_t)$}
\STATE{$Prob_{action} \gets Spectrogram (Prob_{seq})$}
\STATE{$Spec_{real} \gets Spectrogram (A_N)$}

\FOR {number of training iterations} 
\STATE{$Spec_{gen} \gets BigGAN_G (z, Prob_{class},Prob_{seq})$}
\STATE{$R \gets BigGAN_D (Spec_{real}, Spec_{gen})$}
\STATE {Calculate $L_G$ and $L_D$}
\STATE {Update $BigGAN_G$ and $BigGAN_D$}
\ENDFOR

\STATE{$A_{gen} \gets ISTFT(Spec_{gen})$}

\end{algorithmic}
\end{algorithm}

\section{Experimental details}

\subsection{Dataset}
In the context of generating artificial foley tracks from silent video we propose Automatic Foley Dataset (AFD) in our previous work \cite{9126216} that is carefully prepared to avoid external noise focusing on popular foley categories. Since GAN training requires large set of training samples for improved learning, we expand our dataset with more diverse video samples to be used into FoleyGAN training. In Fig. 5 we show the data percentages of individual classes of our updated AFD dataset. The total number video samples are 27800 (of 3 sec duration each). In addition, as an ablation analysis we compare the generated audio sample performance (Table II) by training the proposed FoleyGAN architecture with a subset of AudioSet \cite{gemmeke2017audio} and YouTube8M 
\cite{abu2016youtube}  dataset as these datasets closely complies with our data requirements for this task. We prepare the subsets by collecting videos of similar 12 categories contained in AFD. In all cases, our training set comprises of 80\% and testing set comprises the rest 20\% of the whole datasets.

\begin{center}
\graphicspath{ {Figure} }
\begin{figure}[t] 
\fcolorbox{black}{white}{\includegraphics[width=83mm]{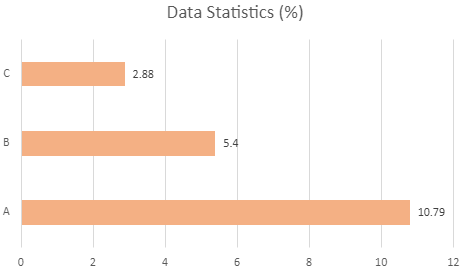}}
\caption{AutoFoley Dataset Statistics: Group A includes car racing, clock ticking, fire rainfall, thundering, typing and waterfall videos; Group B includes chopping, footsteps, gunshots and horse running videos; Group C includes breaking videos.}
\end{figure}
\end{center}

\subsection{Experimental Protocols}

We train event class prediction MTRN network, the consecutive action prediction 2TRN network and the sound generating GAN network separately on training dataset. We collect image features from the output of the $conv5$ layer of the ResNet-50 network. The TRN models have two layers of MLP (256 units in each) for $g_\theta$ and a single layer MLP (12 units) for $h_\phi$. The training for 100 epochs is completed in less than 24 hours on a NVIDIA Tesla V100 GPU. We use minibatch gradient descent with the Adam optimizer \cite{kingma2014adam}. The minibatch size is 128 and learning rate is 0.001.

To implement our audio generation network, we adopt the 512 × 512 BigGAN \cite{brock2018large} architecture (which is a Self-Attention GAN \cite{zhang2019self} based model) trained on our AutoFoley spectral data. In most cases, we follow the similar hyper-parameters and optimisation techniques for the discriminator and generator while training. The whole implementation is done using TensorFlow. Likewise BigGAN, we apply orthogonal Initialization \cite{saxe2013exact} strategy (e.g. introducing a random orthogonal matrix weight in each layer maintaining their orthogonal property) on both the generator and the discriminator. The generator model use skip-z technique to directly link the input latent vector z to specific layers deep in the network where the full dimensionality of z is set to 160 for 512 × 512 spectrogram image generation. We set the learning rate to $2$ × $10^{−4}$ and $5$ × $10^{−5}$ for discriminator and generator respectively. We obey the truncation trick \cite{brock2018large} by resampling the z values to arbitrate between image quality and variety. The overall model is trained via calculating the hinge loss. We use the Adam optimiser \cite{kingma2014adam} for optimisation. BigGAN performance greatly depends on increasing the batch size, more particularly BigGAN requires high batch size training to provide better gradient information while updating the weights through training epochs. However training with larger batch requires GPUs of higher memories. To handle the memory constraints, we implement gradient accumulation technique during our training session. We train our sound generation network  on a single NVIDIA Tesla V100 GPU of 32GB VRAM. Our intuition is to train with a total batch size of 2048. To avail this large batch size without facing "OOM" e.g. out of memory error, we use mini batch size of 128 for 16 gradient accumulations. Our each training session take 8 days to complete for 500 epochs with 12k iterations. We add post-processing filter of 512 length to the generator output for lowering the noisy artifacts of generated spectrogram samples.

\section{Model Evaluation}
In this section, first we describe different numerical evaluation metrices adopted to asses the performance of our proposed method in a quantitative manner and explain the calculated results comparing with state-of-the-art models (subsection (A-E) ). Next, in subsection F we show phase coherence study. Later in subsection G, we present a human survey results to evaluate the generated sound quality on comprehensive way in accordance with the video clips. 

\subsection{Sound Retrieval Accuracy}

We prepare a sound classifier by training a ResNet-50 \cite{He_2016_CVPR} CNN model with spectrogram images of AFD training data. Next, we measure the prediction accuracy of our generated spectrogram samples. We also calculate the classifier's performance by testing it with AFD test spectrogram samples. The average accuracy is measured over all event classes (shown in Table I).

\subsection{Inception Score (IS)}

To evaluate the semantic diversity of generated samples we calculate the inception score (IS) proposed in \cite{salimans2016improved} using the following equation:
\begin{equation}
exp(\mathbb{E}_xD_{KL}(P(y|x)||P(y))   
\end{equation}
Here, $P(y|x)$ represents the conditional class distribution for image sample $x$ predicted by the Inception Network \cite{szegedy2015going} and $P(y)$ gives the marginal class distribution. The equation compute IS score by calculating the Kullback-Leibler (KL) Divergence between these two distributions. The Inception features are extracted from Inception Network \cite{szegedy2015going} trained on the ImageNet dataset. A high IS value is preferred in case of evaluating good generation quality. Since Inception Score evaluation matches with human judgements at a great level, we want to evaluate our generated spectrograms on this basis. Therefore we use our pretrained sound retrieval CNN classifier (mentioned in previous subsection) features to compute the score (shown in Table I and II).

\subsection{Fréchet Inception Distance (FID)}
The Fréchet Inception Distance (FID) measures the Fréchet Distance (FID) between two multivariate Gaussian distributions for synthesized and real samples configuring the mean and covariance of intermediate layer inception features as follows:

\begin{equation}
\begin{split}
FID(r,g) = ||\mu_r - \mu_g||^2 + \\
Tr(\sum r + \sum g - 2(\sum r \sum g)^{1/2})
\end{split}
\end{equation}

here $\mu_r$ and $ \sum r$ represents the mean and covariance of real samples respectively. Likewise, $\mu_g$ and $ \sum g$ represents the mean and covariance of generated of the same. FID score is considered as a good evaluation metric to compare between real data and generated outputs. A low FID score is preferred in case of evaluating good generation quality. Here again we use the same sound retrieval CNN classifier pretrained on AutoFoley spectrograms to compute the FID scores (shown in Table I and II) since existing Inception features pretrained with Imagenet or S09 data will not match our requirements for our specific audio spectrogram generation associated with the video clip.

\subsection{Number of Statistically-Different Bins (NDB)}

We follow another effective quantitative evaluation metric called number of statistically-different bins (NDB) proposed in \cite{richardson2018gans} that takes up two sets of samples from the same distribution and indicates the number of samples that fall into a given bin should be the same up to sampling noise. On other words, NDB score shows the number of cells where the training sample number is statistically different from generated sample number through a two-sample binomial test. Here we cluster our train samples into $k = 50$ Voronoi cells into log-spectrogram by $k$ -means clustering. Next, we assign the generated samples to the nearest cell by mapping them into the log-spectrogram space. Certainly a low NDB score is preferred in case of evaluating good generation quality. Table I and II show NDB scores for different sound generative models on AFD data as well as NDB scores for FoleyGAN model trained on different datasets respectively. In addition to analyze generated sample quality for individual classes with different sound encoding method we compute the scores and present the ablation study in Table III.

\subsection{Quantitative Study Result Analysis}
We perform quantitative experiments (mentioned in above subsections) on the generated samples from our proposed FoleyGAN and other baseline audio generating networks and presented the results in Table I, where all the models are trained on our AFD dataset. The FoleyGAN model with visual guidance acheives the highest IS score (10.97) and sound retrieval accuracy (76.08\%) that are very close to the experiment results with the real samples. However, the generated sample performance performance deteriorates (lower than AutoFoley and GANSYNTH samples) when FoleyGAN is not guided with visual action information. The same trend follows in case of FID and NDB computations. Our proposed FoleyGAN with visual guidance results lowest (better) scores (67 and 18.47 for FID and NDB respectively) which again represents good generation quality. Next, we want to evaluate our proposed model efficiency on two most popular video dataset (YouTube8M, AudioSet), comparative results are shown in Table II. Since most of the audio clips associated with YouTube8M and AudioSet video samples consist background noise and sometimes sounds from multiple sources, it somewhere becomes difficult for the generator to learn the original pattern from latent $z$ from similar number of training epochs used with AFD video samples. However, the scores are not too far from real data that leads to the fact that despite foley generation our proposed model can be deployed in generalized applications of audio synthesis into silent video inputs as well. Later in Table III, we present NDB scores of generated samples of individual AFD class on FoleyGAN models using 5 different sound encoding (eg. Short-Time-Fourier Transform (STFT), Mel-Spectrum (MS), Mel-Frequency Cepstral Coefficient (MFCC), Log-amplitude of Mel-Spectrum (LMS) and Constant-Q Transform (CQT) ) as GAN inputs. All class results show the lowest value of NDB is calculated for the generated samples where FoleyGAN is trained with LMS audio features. 

\captionsetup[table]{name=TABLE,labelsep=newline,textfont=sc}
\begin{table*}[!h]
\begin{center}
\caption{\textcolor{Black}{Performance Comparison of Generated Samples\\
from Sound Generative Baseline Models with AFD Dataset}} 
\begin{tabular}{|c|c|c|c|c|}
\hline
\textbf{Samples} & \textbf{IS} & \textbf{FID} & \textbf{NDB} & \textbf{Average Accuracy (\%)} \\ \hline 
Real Data & 11.42 & 11 & 3.23 & 78.32\\\hline
\textbf{FoleyGAN} & \textbf{10.97} & \textbf{67} & \textbf{18.47} & \textbf{76.08}\\\hline
FoleyGAN without visual guidance & 9.22 & 181 & 26.53 & 64.61\\\hline
AutoFoley (Frame Sequence Network) & 10.40 & 127 & 20.94 & 65.79\\\hline
AutoFoley (Frame Relation Network) & 10.72 & 119 & 20.03 & 63.40\\\hline
GANSYNTH (IF-Mel + H) & 10.87 & 115 & 22.14 & 73.12\\\hline
SpecGAN & 8.62 & 271 & 30.07 & 61.75\\\hline
WaveGAN & 7.36 & 322 & 34.91 & 59.93\\\hline
\end{tabular}
\end{center}
\end{table*}

\captionsetup[table]{name=TABLE,labelsep=newline,textfont=sc}
\begin{table*}[!h]
\begin{center}
\caption{\textcolor{Black}{Performance Comparison of Generated Samples\\
from FoleyGAN with Audio-Visual Datasets}} 
\begin{tabular}{|c|c|c|c|c|}
\hline
\textbf{Dataset} & \textbf{IS} & \textbf{FID} & \textbf{NDB} & \textbf{Average Accuracy (\%)} \\ \hline 
Real Data & 11.42 & 11 & 3.23 & 78.32\\\hline
\textbf{AFD} & \textbf{10.97} & \textbf{67} & \textbf{18.56} & \textbf{76.08}\\\hline
YouTube8M Subset & 10.04 & 114 & 20.03 & 70.16\\\hline
AudioSet Subset & 9.72 & 102 & 21.16 & 68.71\\
\hline
\end{tabular}
\end{center}
\end{table*}

\begin{table}[!h]
\begin{center}
\caption{\textcolor{Black}{Generated Sample Quality Comparison with AFD dataset for different sound feautures using FoleyGAN}}
\begin{tabular}{| c | c | c | c | c| c |}
\hline
\textbf{Class} & \multicolumn{5}{ c |}{\textbf{NDB (k = 50)}} \\\cline{2-6} & \textbf{STFT} & \textbf{CQT} & \textbf{MS} & \textbf{MFCC} & \textbf{LMS} \\ \hline
Break &  31.6 & 30.1 & 28.4 & 29.3 & \textbf{23.5} \\ \hline
Car & 22.8 & 27.5 & 30.2 & 21.8 & \textbf{21.6} \\ \hline
Clock & 15.3 & 20.4 & 19.1 & 14.0 & \textbf{11.2}  \\ \hline
Chopping & 21.6 & 18.5 & 22.1 & 17.2 & \textbf{15.7}  \\ \hline
Fire & 15.1 & 17.4 & 15.3 & 13.6 & \textbf{12.0}\\ \hline
Footstep & 18.9 & 21.0 & 19.1 & 18.2 & \textbf{13.3} \\ \hline
Gunshot & 26.7 & 28.1 & 30.4 & 25.6 & \textbf{24.5}\\ \hline
Horse & 19.9 & 20.3 & 21.2 & 19.1 & \textbf{17.3}\\ \hline
Rain & 12.8 & 13.4 & 13.9 & 12.4 & \textbf{12.1}\\ \hline
Thunder & 34.1 & 31.7 & 36.5 & 35.3 & \textbf{33.8}\\ \hline
Typing & 27.6 & 29.8 & 31.0 & 28.1 & \textbf{27.2}\\ \hline
Waterfall & 10.7 & 11.5 & 12.3 & 11.2 & \textbf{9.4} \\ \hline
\textbf{Average} & 21.43 & 22.48 & 23.29 & 20.48 & \textbf{18.47}\\ \hline
\end{tabular}
\end{center}
\end{table}

\subsection{Phase Coherence}

To envision the phase coherence between training and generated waveforms, we show Rainbowgram representations \cite{engel2019gansynth} of each event class in Fig.6 where the left column is indicating rainbowgrams of originals and the right one is displaying the same of generated tracks. The comparison between two rainbowgrams helps to visualize both the phase consistency and differences of the wave harmonics in a more clear way. In every rainbowgram image, the brightness symbolizes the log magnitudes and the color depicts the instantaneous frequencies of the respective waveform. Noticeably the rainbowgrams of fire, footstep, rain, waterfall class synthesized waves depict vigorous consistent colors and phase coherence like the same of real waves. Few deformities in color lines are noticed in breaking, chopping, ticking clock, running horse, typing, thundering categories. However, rainbowgrams of car, gunshot classes show more phase discontinuities since the wave harmonics are occasionally afflicted by noise components which are responsible for the additional color flecks, phase irregularities and aperiodicities.

\begin{figure*}[]
\centering
\subfloat{
  \includegraphics[height=7.7cm,width=.74\textwidth]{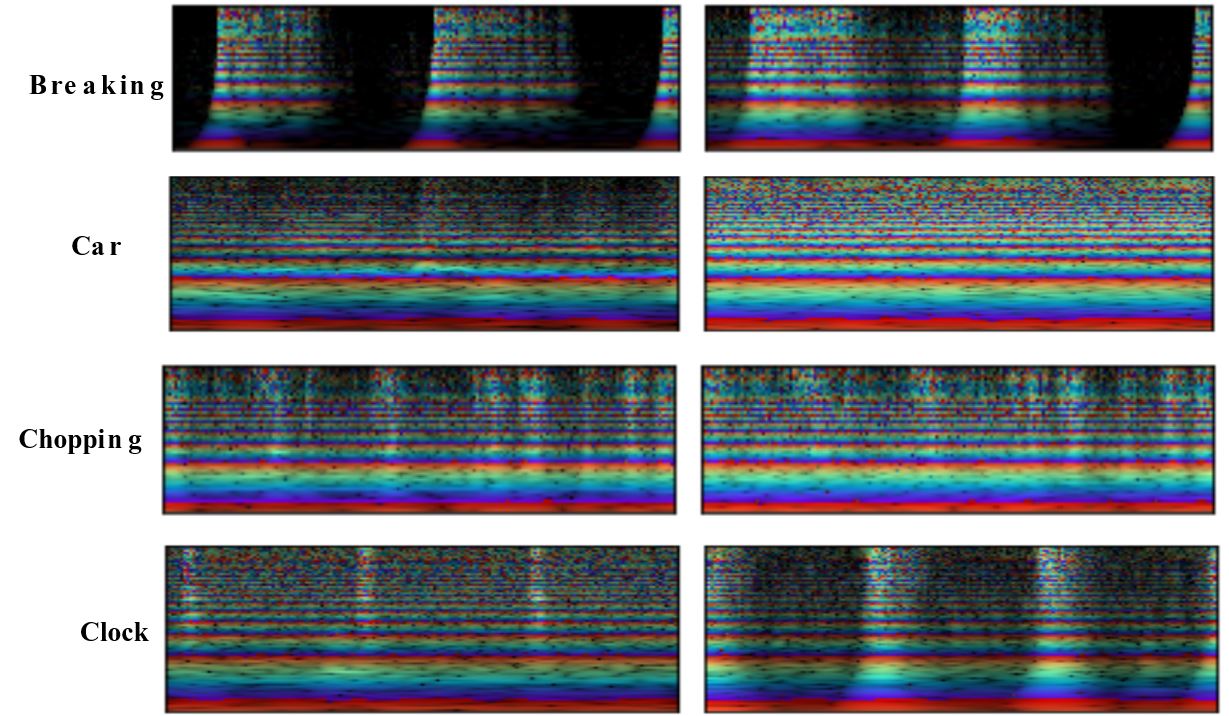}
}\\
\subfloat{
  \includegraphics[height=7.9cm,width=.74\textwidth]{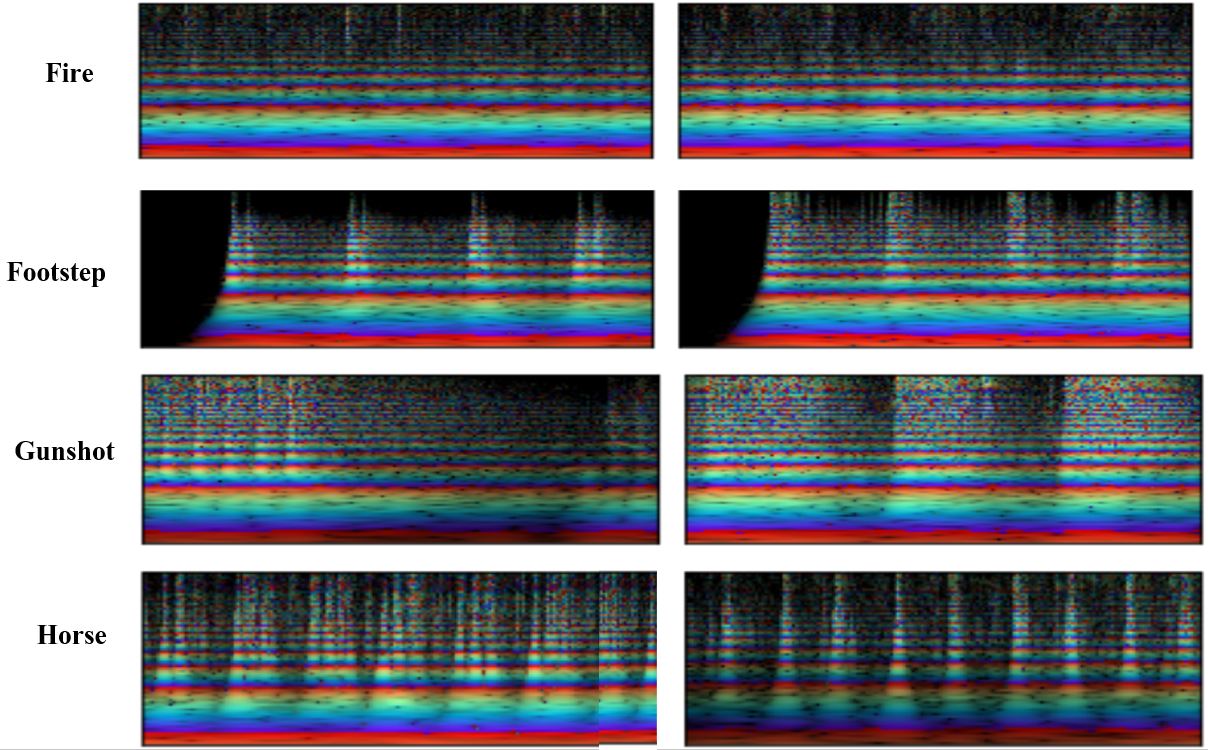}
}\\
\subfloat{
  \includegraphics[height=7.9cm, width=.74\textwidth]{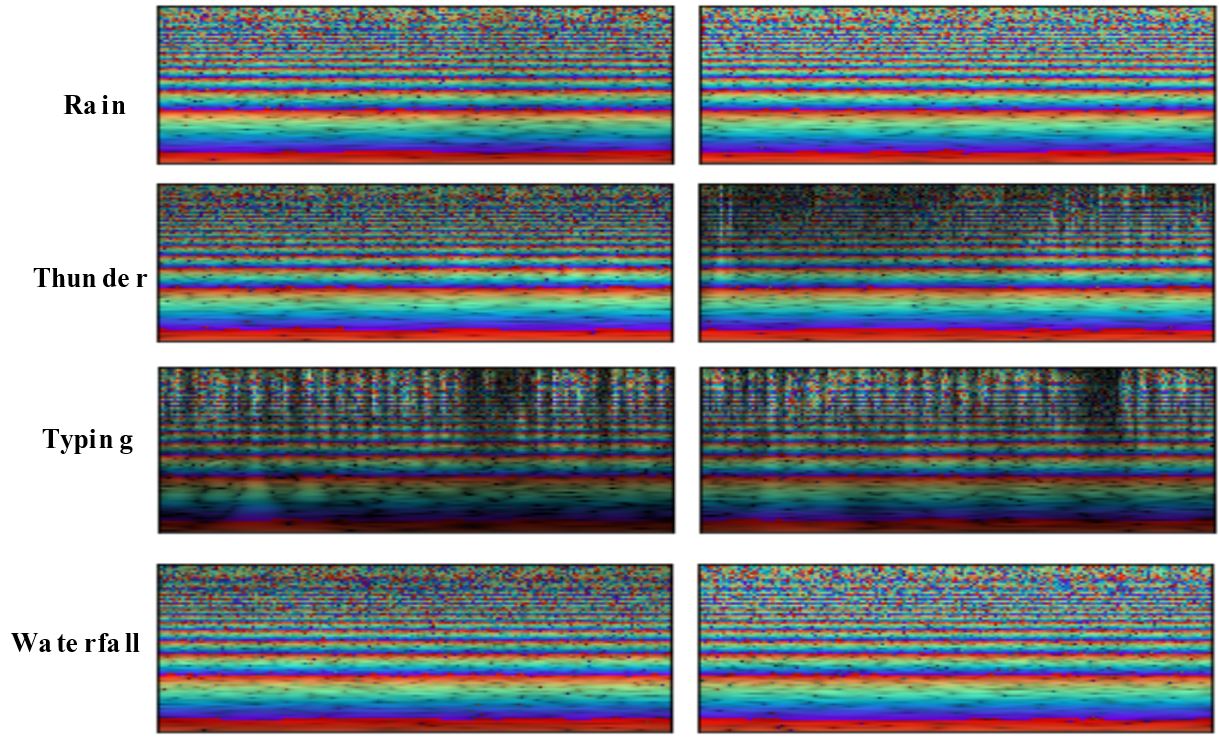}
}

\caption{Phase coherence comparison between original and generated sound samples through Rainbowgram resprentation. Horizontal and vertical axes are showing time and frequency respectively. }
\end{figure*}

\subsection{Qualitative Study: Human Survey}

We find human survey is an inevitable assessment to judge both the audio quality and its synchronicity with the video recording, since human brain can inherently perceive the correspondence between audio-visual modalities in such coinciding events. Therefore, we prepare a research study participated by our college of engineering students and officials to survey qualitative questions on our synthesized sound tracks superimposed on the real video clips. There we set 2 queries for videos of each event class. Every audience is asked to observe videos with our synthesized sounds and rate the generated sample on the basis of the overall quality of the audio (question 1) and how much they perceive the audio is synchronous with the visual scene (question 2). The observance score is marked out of a scale of 10. This experiment is conducted over 100 participants. Through this approach we intent to capture human's natural intuition to asses the artificially synthesized sound quality so that we can determine the level of our generated sound traits that signifies how much it is capable to portray the original event. 

Table IV presents the average ratings for individual class separately on both queries. The best result for both queries comes for the waterfall sound. We think it is because the training sound clips contain a similar continuous pattern for this category and thus generator learns it more accurately. Regarding audio-visual synchronicity, rainfall (9.6) and fire (9.4), clock ticking (9.2) event classes are other three classes to capture the continuous pattern of sound. Additionally asynchronous event classes e.g. chopping on kitchen board (9.5), footstep (9.3), horse running (9.2), breaking (9.0) and car (8.8) also provide outstanding syncing score. Since object movements are more visible due to close up video recordings ( mostly in chopping, footstep and breaking videos) we assume that is assisting towards generating more synchronous sound with visual guidance. However, in generated horse clips we find some variation in sound intensity when the horse is hitting the ground while running or walking. Depending on the action speed the sound intensity and pitch change which is a challenging property to learn. In few cases, we observe the model is unable to capture this trait and it is learned to generate more general form of horse running sound. Despite that, the generated tracks are well synced with the test clip indicating the success in visual guidance introduced to the GAN.

For gunshot and thundering videos, we have to rely on videos that are available online for use as are not able to record them in person. The thundering category is the most challenging part, in most cases the lightening visuals were unable to provide action info coherently with audio features while generating the sound. However, if we consider the audio quality the generated thundering audio clip sounds similar to the raining sound. In case of gunshot sound generation we find the action of shoots are not clearly visible in most of the recordings due to distant object placement. This may hinder providing temporal action updates to the GAN while generating the sound.  

We have the least number of training examples in breaking category (mostly collected from online sources). We assume this class needs to be developed with more training samples with inclusion of variety of object materials to expect better audio quality. According to this study people perceive ticking clock, footstep, fire, running horse, rain and water audio quality really well; car, chopping, gunshot, typing sound as on average and thundering sound as the least similar to originals. Averaging all class results, our generated sound score 7.1 and 8.1 out of 10 in terms of quality and synchronicity with video respectively.

\begin{table}[!h]
\begin{center}
\caption{\textcolor{Black}{Human Evaluation Results}}
\begin{tabular}{| c | c | c |}
\hline
\textbf{Class} & \textbf{Audio Quality} & \textbf{Audio-Visual Synchronicity} \\ \hline
Break &  7.7 & 9.0 \\ \hline
Car & 5.1 & 8.8\\ \hline
Clock & 7.5 & 9.2 \\ \hline
Chopping & 5.6 & 9.5  \\ \hline
Fire & 8.2 & 9.4\\ \hline
Footstep & 9.1 & 9.3 \\ \hline
Gunshot & 5.9 & 4.3\\ \hline
Horse & 8.4 & 9.2\\ \hline
Rain & 8.9 & 9.6\\ \hline
Thunder & 3.2 & 3.6\\ \hline
Typing & 6.3 & 5.5\\ \hline
Waterfall & 9.2 & 9.8\\ \hline
\textbf{Average} & \textbf{7.1} & \textbf{8.1}\\ \hline
\end{tabular}
\end{center}
\end{table}

\section{Conclusion and Future Scope}
In this paper, we address the time synchronization setback in the task of visual to audio generation and take the first attempt to exploit conditional GANs with visual guidance of an event to synthesize visually aligned sound. For efficient adversarial training we expand the AFD dataset with adequate diverse video samples in each class. In order to evaluate our models, we conduct numerical and qualitative evaluations and compared with baseline models with leading results. Our experiments reveal successful synchronous sound synthesis capability of our proposed FoleyGAN system maintaining good audio quality that can indeed be used as automatic Foley generators for silent movie scenes as well as for other audio-visual intersensory applications. One shortcoming in this work is the requirement that the subject of classification is present in the entire video frame sequence. Furthermore, in our approach we have not dealt with video clips containing multiple sound sources. In addition we want to work with more sound categories. These are the targeted directions of our future work.

\appendices





\bibliographystyle{IEEEtran}
\bibliography{FoleyGAN.bib}{}
\begin{IEEEbiography}[{\includegraphics[width=1in,height=1.25in,clip,keepaspectratio]{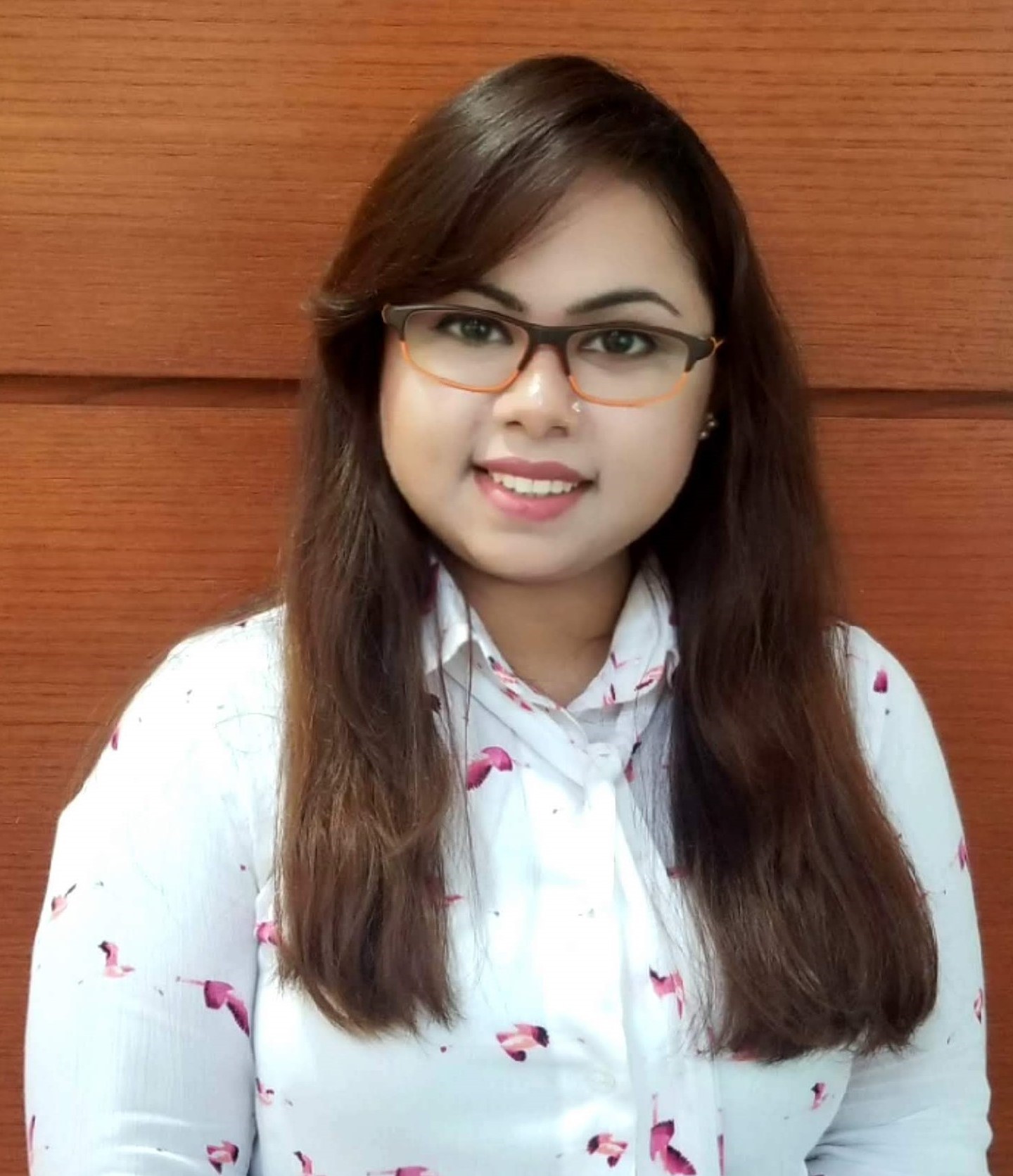}}]{Sanchita Ghose} (S'20)
received a bachelor’s degree in electrical and electronic engineering from Ahsanullah University of Science and Technology (AUST), Dhaka, Bangladesh in 2013. She is currently working toward a doctoral degree with the Cloud Lab for Engineering Application and Research (CLEAR), at the University of Texas at San Antonio, Texas, USA. She is a member of the Open Cloud Institute (OCI), IEEE, and Society of Women Engineers (SWE). She had previously published in IEEE Transaction on Multimedia and had been reviewer of the same and IEEE Access. Her current research interest includes developing a deep learning algorithm for multimodal learning and cross-modal retrieval applications, focusing on computer vision, action recognition, sound synthesis and video processing.
\end{IEEEbiography}
\begin{IEEEbiography}[{\includegraphics[width=1in,height=1.25in,clip,keepaspectratio]{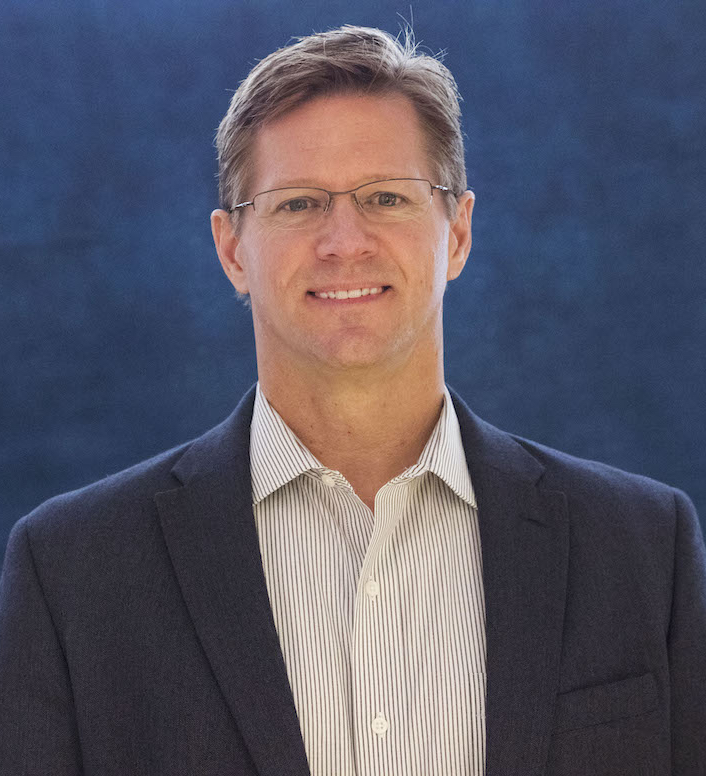}}]{John J. Prevost} (S'06-M'13-SM'20)
received his first B.S. degree from Texas A\&M in economics in 1990. He received his second B.S. degree in electrical engineering from the University of Texas at San Antonio (UTSA), where he graduated magna cum laude in December 2009. In 2012 he received his M.S. degree in Electrical Engineering, also from UTSA along the way to earning his Ph.D. in electrical engineering in December 2013. His current academic appointment is Assistant Professor in the Department of Electrical and Computer Engineering at UTSA. In 2015, he cofounded and became the Chief Research Officer and Assistant Director of the Open Cloud Institute. Prior to his academic appointment, he has served as Director of Product Development, Director of Information Systems and Chief Technical Officer for various technical firms. He remains an active consultant in areas of complex systems and cloud computing and maintains strong ties with industry leaders. His current research interests include energy-aware cloud optimization, cloud-controlled robotics, cloud-based communications, and quantum cloud computing.
\end{IEEEbiography}
\end{document}